\documentclass[letterpaper]{article} 
\usepackage{aaai2027}  
\usepackage{amsmath}
\usepackage{amssymb}
\usepackage[utf8]{inputenc}
\DeclareMathOperator{\KL}{D_{\mathrm{KL}}}
\usepackage[hyphens]{url}  
\usepackage{graphicx} 
\usepackage{natbib}  
\usepackage{caption} 
\usepackage{algorithm}
\usepackage{algorithmic}
\usepackage{xspace}
\usepackage{colortbl}

\newcommand{\mypara}[1]{\smallskip\noindent{\bf {#1}.}\xspace}

\newcommand{\sysname}{PAMA\xspace}
\usepackage{newfloat}
\usepackage{listings}
\usepackage{multirow} 
\DeclareCaptionStyle{ruled}{labelfont=normalfont,labelsep=colon,strut=off} 
\floatstyle{ruled}
\newfloat{listing}{tb}{lst}{}
\floatname{listing}{Listing}
\usepackage{svg}
\usepackage{booktabs}
\let\titleold\title
\renewcommand{\title}[1]{\titleold{#1}\newcommand{\thetitle}{#1}}
\def\maketitlesupplementary
   {
   \newpage
       \twocolumn[
        \centering
        \Large
        \textbf{\thetitle} \\
        Supplementary Material \\
        \vspace{1cm}
       ] 
   }

\title{Policy Alignment: New Signals for Membership Auditing in On-Policy Distillation}
\author{
    Yilong Yang\textsuperscript{\rm 1}, WenZhuo Shang\textsuperscript{\rm 1}, Yule Liu\textsuperscript{\rm 2}\corresponding, Jiale Teng\textsuperscript{\rm 1}, Zhuo Ma\textsuperscript{\rm 1}
}
\affiliations{
    \textsuperscript{\rm 1}Xidian University \quad \textsuperscript{\rm 2}Hong Kong University of Science and Technology (Guangzhou)

}

\begin{document}

\maketitle

\begin{abstract}
On-policy distillation (OPD) trains a student model by aligning its policy with a teacher model on trajectories generated by the student model itself.
Through this process, the student policy moves toward the teacher on the prompts used for distillation.
However, these prompts are often private and costly, creating a need for prompt-level membership auditing.
Existing methods mainly rely on likelihood-based confidence signals or student policy drift between checkpoints, but they do not capture the teacher-induced direction of the student update.
In this paper, we propose Policy Alignment Membership Auditing (\sysname), a new auditing framework tailored for OPD.
Our key observation is that a member prompt directly contributes to the teacher-guided policy update, while a non-member prompt only experiences indirect effects through cross-prompt generalization.
Based on this directional trace, \sysname measures whether the student update moves toward reducing the teacher loss on a candidate prompt.
Specifically, we introduce Teacher Alignment Gain (TAG) to estimate the teacher-aligned update direction from model outputs, and further combine it with student drift and uncertainty alignment signals for reliable membership auditing.
We evaluate \sysname on six datasets and three teacher--student model families.
On MATH, the primary evaluation benchmark, \sysname achieves AUC values of 0.791--0.941, improving AUC by 14.6--20.6\% over state-of-the-art baselines.
\end{abstract}

\begin{figure}[t]
    \centering
    \includegraphics[width=\columnwidth]{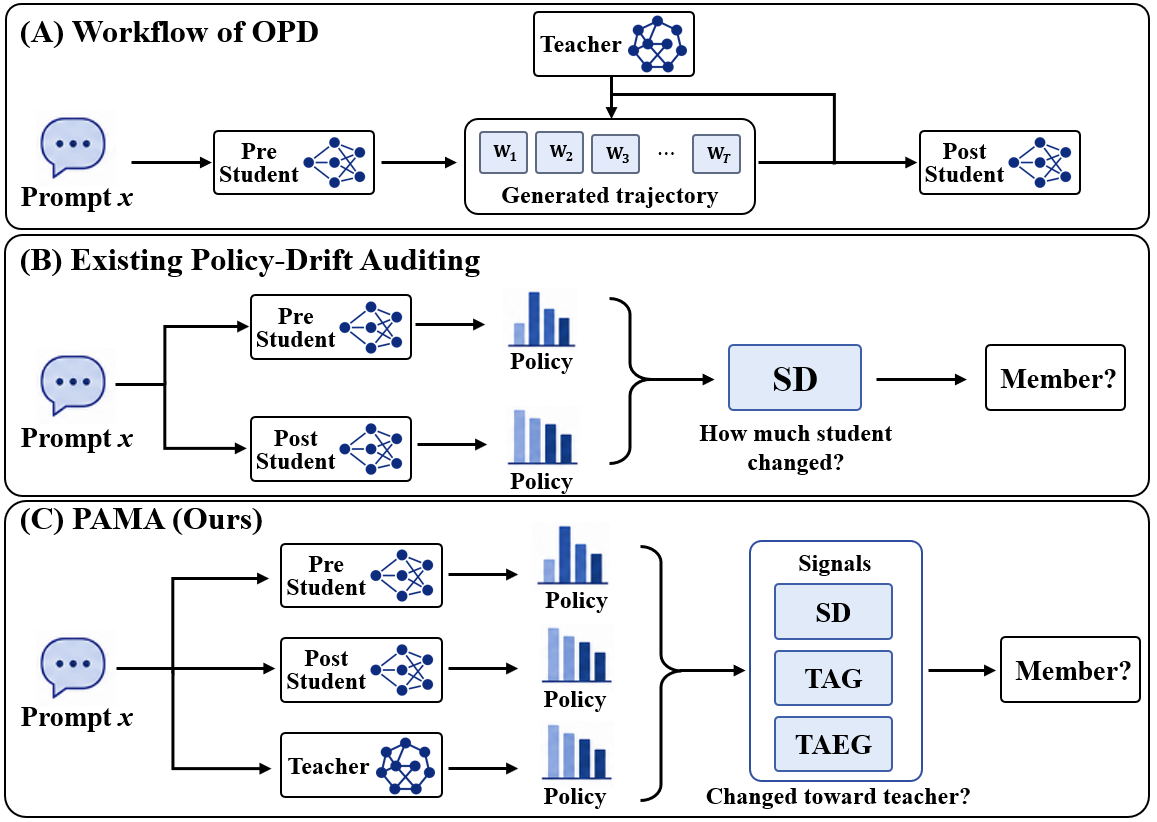}
    \caption{Overview of OPD prompt and membership auditing:
    (A) teacher-guided student training, (B) policy-drift auditing, and
    (C) \sysname with teacher-directed alignment.}
    \label{fig:intro}
\end{figure}

\section{Introduction}

On-policy distillation (OPD) has recently emerged as an effective paradigm for improving the capabilities of student models by learning from stronger teacher models~\cite{xiao2026mimo,zeng2026glm,zhao2026self,hubotter2026reinforcement}.
Unlike traditional distillation methods that rely on fixed training samples, OPD allows the student model to generate its own trajectories and gradually align its policy with the teacher model.
During this process, the teacher-guided update direction plays a critical role, as it determines how the student policy evolves toward the teacher on the prompts used for distillation.

However, the prompts used in OPD are often private, sensitive, and valuable, as they may contain proprietary data or carefully designed reasoning tasks.
Unauthorized disclosure of these prompts may expose confidential information and compromise the ownership of valuable training resources.
Therefore, prompt-level membership auditing~\cite{shokri2017membership,liu2026diba}, which aims to determine whether a given prompt was used during OPD training, becomes an important and practical security requirement.

Existing membership auditing methods are mainly designed for conventional supervised training scenarios and cannot fully capture the unique characteristics of OPD.
Likelihood-based methods~\cite{duan2024membership,puerto2025scaling,zhang2025min} usually require predefined target responses or rely on output confidence, which are unavailable for the self-generated trajectories in OPD.
Another line of methods~\cite{liu2026diba} measures student policy drift between checkpoints, but only focuses on the magnitude of model changes and ignores whether these changes are guided by the teacher model.
Although recent studies reveal the complex dynamics of on-policy updates~\cite{fu2026revisiting,yang2026learning}, such drift only reflects \textit{how much} the student changes, while leaving \textit{where the student moves} unresolved.

In this paper, we identify a new phenomenon in OPD: a member prompt directly participates in the teacher-guided policy update, causing the student policy to move toward reducing the teacher loss on this prompt.
In contrast, a non-member prompt is only affected indirectly through cross-prompt generalization, resulting in a weaker teacher-aligned update signal.
Based on this observation, we propose Policy Alignment Membership Auditing (\sysname), a framework tailored for prompt-level membership auditing in OPD.
\sysname introduces Teacher Alignment Gain (TAG) to quantify the extent to which the student update aligns with teacher guidance on a candidate prompt.
To estimate TAG without accessing training gradients, we design a two-point zeroth-order approximation that measures the change in teacher loss before and after distillation on the same student-generated trajectories, providing an efficient estimation of the teacher-aligned update direction.
Furthermore, \sysname integrates TAG with complementary signals, including Zlib, Min-K, student drift and teacher-aligned entropy gain signals, to achieve reliable membership auditing.

We evaluate \sysname on six datasets and three teacher--student model families under two practical access settings: Full-distribution access and Top-$K$ access to teacher model outputs.
Compared with SOTA baselines, \sysname improves AUC by 0.101--0.146 across different model families, corresponding to relative improvements of 14.6--20.6\%.



Our main contributions are:
\begin{itemize}
\item We identify teacher-guided policy alignment as a new membership signal in OPD and reveal that member prompts directly drive teacher-aligned updates, while non-member prompts only exhibit indirect effects through cross-prompt generalization.

 \item We propose \sysname, which estimates this alignment signal through a Teacher Alignment Gain, and further integrates TAG with student drift and teacher-aligned entropy gain signals to enable reliable auditing under different teacher-access settings.
 
 \item We evaluate \sysname across six datasets and three teacher--student model families, validate the effectiveness of TAG as an approximation of teacher-aligned updates, and demonstrate its advantages over existing methods under both Full-distribution and Top-$K$ access.
\end{itemize}

\section{Preliminaries \& Threat Model}
\label{sec:background_threat}

\subsection{On-Policy Distillation}
\label{sec:opd_background}

OPD trains a student to match a fixed teacher $T$ on responses generated by
the student itself~\cite{agarwal2024policy,gu2024minillm}.  $x$ is a prompt,
$y=(y_1,\ldots,y_{L(y)})$ a generated response of length $L(y)$, and
$t\in\{1,\ldots,L(y)\}$ a token position. The prefix before position $t$ is
$y_{<t}=(y_1,\ldots,y_{t-1})$, and $\mathcal V$ denotes the token vocabulary.

We distinguish the pre-distillation student $S_{\mathrm{pre}}$, the post-distillation student $S_{\mathrm{post}}$, and the teacher $T$. 
For $j\in\{\mathrm{pre},\mathrm{post},T\}$ and a candidate next token
$v\in\mathcal V$, $\pi_j(v\mid x,y_{<t})$ is the probability assigned to $v$
by the corresponding model. During OPD, $\pi_\theta$ denotes the current
student policy with trainable parameters $\theta$. Let $\mathcal D_{\mathrm{OPD}}$ be the set of prompts used
for distillation.

Given $x\sim\mathcal D_{\mathrm{OPD}}$, the current student samples
$y\sim\pi_\theta(\cdot\mid x)$. Let $\KL(P\Vert Q)$ denote the KL divergence
from distribution $P$ to distribution $Q$. A representative reverse-KL OPD
objective averages the token-level teacher--student divergence along the
sampled response:
\begin{equation}
\label{eq:opd_objective}
\mathcal L_{\mathrm{OPD}}=
\mathbb E_{x,y}\!\left[\frac{1}{L(y)}\sum_{t=1}^{L(y)}
\KL\!\left(\pi_\theta\Vert\pi_T\right)\right].
\end{equation}
Here $\mathbb E_{x,y}$ is taken over the two sampling steps above, and every
KL term is conditioned on the same $(x,y_{<t})$; this common condition is
omitted in Eq.~\ref{eq:opd_objective} to keep the formula on one line.

\subsection{Membership Auditing}
\label{sec:mia_background}

Membership auditing answers whether a candidate data point belongs to the training set of a target model~\cite{shokri2017membership}.
For language models, common attacks score an observed token sequence using its likelihood, compressed length, or least-likely tokens~\cite{duan2024membership,carlini2021extracting,shi2023detecting,zhang2025min,mattern2023membership,xie2024recall,fu2023practical}.
These methods assume that the audited example includes the target tokens whose probability is evaluated.
In OPD, however, membership belongs to a prompt, and the student samples a new response whenever that prompt is used.
Applying a text-level score to an audit-time response therefore measures confidence in that sampled response rather than directly measuring whether the prompt was used for distillation.
DIBA instead audits reinforcement-learning data through changes between policy checkpoints~\cite{liu2026diba}.
Its policy-shift signal transfers to OPD, while its reward- and optimization-dependent signals require training information that an external OPD auditor does not observe.

\subsection{Threat Model}
\label{sec:threat_model}

After a provider completes OPD, an external auditor needs to determine whether a sensitive prompt was used for distillation.
We study this post-hoc setting, in which the auditor neither participates in nor modifies the completed OPD run.

\mypara{Auditor's Goal}
Given a candidate prompt $x$, the auditor predicts whether it was used in the
audited OPD run:
\[
H_1:x\in\mathcal D_{\mathrm{OPD}}
\qquad\text{and}\qquad
H_0:x\notin\mathcal D_{\mathrm{OPD}}.
\]
The membership label of $x$ is unavailable when this prediction is made.

\mypara{Auditor's Capabilities}
In every setting, the auditor has inference-only access to the pre- and post-distillation students.
For any prompt and response prefix, it can query their complete next-token distributions and sample responses.
The interface does not reveal either student's parameters, expose gradients, or permit model updates.
The auditor also has a labeled reference set that is disjoint from the candidate prompts being tested.
This set contains known members from the same audited OPD run and matched non-members.
Following classic supervised attack-model settings, the labels are used to standardize the audit features and fit the predictor~\cite{shokri2017membership,puerto2025scaling,liu2026diba}.
The auditor does not observe the OPD trajectories, optimizer state, or per-prompt training losses and gradients.
The two access settings below describe the teacher reference available at the same prompt and response prefixes:
\begin{itemize}
    \item \textbf{Full-distribution Access.}
    The teacher interface returns
    $\pi_T(\cdot\mid x,y_{<t})$ over the complete vocabulary at every response
    position.
    \item \textbf{Top-$K$ Access.}
    The teacher interface returns only the $K$ highest-probability tokens and
    their probabilities at each position; the remaining probability mass is
    hidden.
\end{itemize}

\section{Method}
\label{sec:method}

Our goal is to infer whether a candidate prompt $x$ directly participated in OPD.
A member prompt directly contributes a teacher-guided gradient to the student update, while the policy on a non-member prompt changes only indirectly through cross-prompt generalization.
We test for this difference using the output distributions of the pre- and post-distillation students and the available teacher reference.

\subsection{Audit Contexts}
\label{sec:audit_notation}

Our scores compare checkpoint distributions at fixed token contexts.
Let $\theta_{\mathrm{pre}}$ and $\theta_{\mathrm{post}}$ denote the student parameters before and after OPD, respectively, and let
$\pi_{\mathrm{pre}}=\pi_{\theta_{\mathrm{pre}}}$ and
$\pi_{\mathrm{post}}=\pi_{\theta_{\mathrm{post}}}$ be their policies.
We use $s\in\{\mathrm{pre},\mathrm{post}\}$ to index these two checkpoints.
Policy $\pi_s$ induces a context distribution $\mu_x^s$ by sampling
$y\sim\pi_s(\cdot\mid x)$ and taking every prefix
$c_t=(x,y_{<t})$.
We draw $R$ responses from this distribution and reuse the resulting contexts
$c_t^{(r)}=(x,y_{<t}^{(r)})$ for every student and teacher evaluation.
For a token-level quantity $f$, its empirical average is
\begin{equation}
\label{eq:prompt_aggregation}
\mathcal A_x^{s}[f]
=
\frac{1}{R}\sum_{r=1}^{R}
\frac{1}{L(y^{(r)})}
\sum_{t=1}^{L(y^{(r)})}f(c_t^{(r)}).
\end{equation}
Within each score, using one realized context set ensures that every student
and teacher distribution is evaluated at the same inputs.
SD uses $\mathcal A_x^{\mathrm{post}}$.
TAG and TAEG use $\mathcal A_x^{\mathrm{pre}}$.

\subsection{Student Drift}
\label{sec:student_drift}

Following DIBA's policy-side divergence signal, we measure how much the
post-distillation policy differs from the pre-distillation checkpoint on the
candidate prompt~\cite{liu2026diba}.
Using the post-student contexts defined above, we compute
\begin{equation}
\label{eq:sd}
s_{\mathrm{SD}}(x)=\mathcal A_x^{\mathrm{post}}\!\left[
\KL\!\left(\pi_{\mathrm{post}}\Vert\pi_{\mathrm{pre}}\right)\right].
\end{equation}
We refer to this score as Student Drift (SD); a larger value indicates a
larger policy change on $x$.
SD records the magnitude of checkpoint change, while the next test examines
whether that change follows the teacher-directed OPD objective.

\subsection{Teacher-Directed Progress}
\label{sec:teacher_alignment}

Student Drift shows how much the policy changed on $x$.
We next determine whether this change follows the teacher-directed objective
that drives OPD.

\mypara{Insight}
OPD trains the student with reverse-KL divergence on student-generated
contexts.
At the initial checkpoint, the pre-student contexts follow the same on-policy
distribution used by this objective.
They therefore define a common reference for asking whether the completed
checkpoint update follows the initial teacher-loss descent direction on $x$.
On these contexts, the prompt-specific teacher discrepancy is
\begin{equation}
\label{eq:teacher_potential}
\mathcal D_x^T(\theta;\mu_x^{\mathrm{pre}})
=\mathcal A_x^{\mathrm{pre}}\!\left[
\KL\!\left(\pi_\theta\Vert\pi_T\right)\right].
\end{equation}
With author-side parameter access, let
$g_x^{\mathrm{pre}}
=\nabla_\theta
\mathcal D_x^T(\theta_{\mathrm{pre}};\mu_x^{\mathrm{pre}})$
and let
$\Delta\theta=\theta_{\mathrm{post}}-\theta_{\mathrm{pre}}$.
The negative gradient $-g_x^{\mathrm{pre}}$ is the local parameter direction
that most rapidly reduces the teacher discrepancy on $x$.
We score the completed checkpoint displacement against this initial direction:
\begin{equation}
\label{eq:oracle_teacher_projection}
q_{\mathrm{proj}}^T(x)
=-\left(g_x^{\mathrm{pre}}\right)^\top\Delta\theta.
\end{equation}
A positive value means that the update is aligned with a direction that
locally decreases the prompt-specific teacher discrepancy.
This is an unnormalized projection, so its magnitude also scales with the norm
of the prompt-specific gradient.

\begin{figure}[t]
    \centering
    \includegraphics[width=\columnwidth]{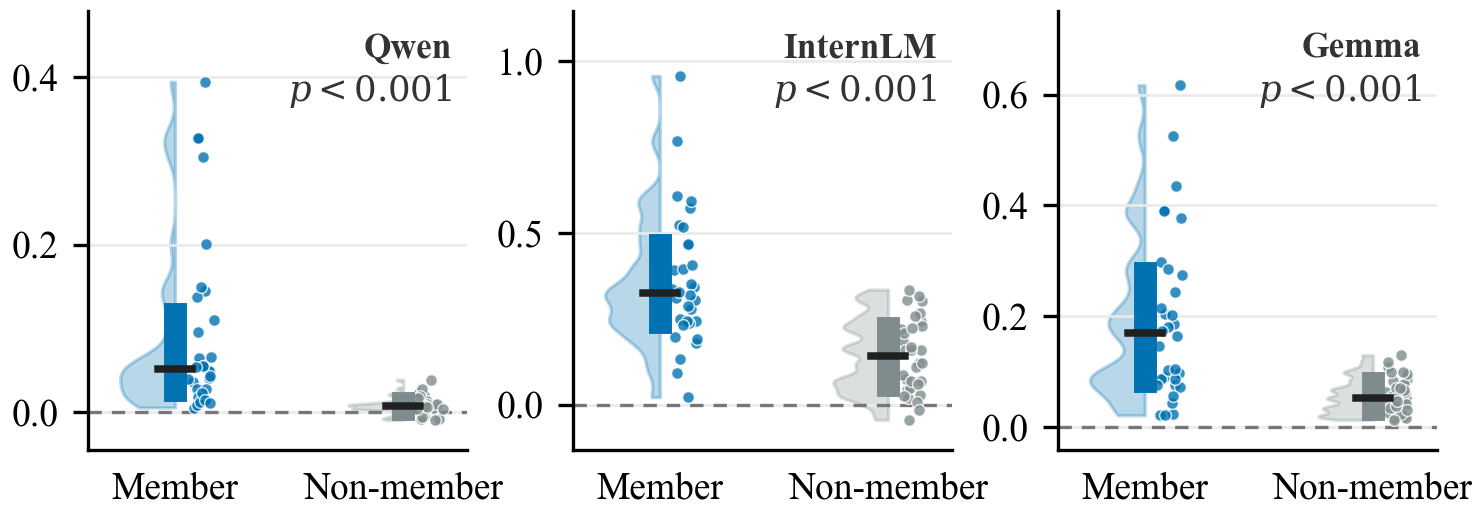}
    \caption{Teacher-directed audit projection for member and non-member prompts.}
    \label{fig:oracle_teacher_projection}
\end{figure}

\mypara{Why Membership Affects the Projection}
Let $\mathcal S_{\mathrm{tr}}$ denote the prompts used for OPD training, all of
which are members, and let $x'$ index a prompt in this set.
At initialization, consider the local full-batch gradient step
$\delta\theta_0
=-\frac{\eta}{|\mathcal S_{\mathrm{tr}}|}
\sum_{x'\in\mathcal S_{\mathrm{tr}}}g_{x'}^{\mathrm{pre}}$, where
$g_{x'}^{\mathrm{pre}}$ is defined in the same way as
$g_x^{\mathrm{pre}}$.
Because every $x'$ belongs to the training set, $\delta\theta_0$ contains only
member gradients.
We use $g_x^{\mathrm{pre}}$ to probe whether this member-only update follows the
teacher-loss descent direction for the candidate $x$; when $x$ is a non-member,
its gradient is evaluated only for this probe and never contributes to
$\delta\theta_0$.
Projecting this step onto the descent direction for the candidate $x$ gives
\begin{equation}
\label{eq:local_membership_decomposition}
\begin{aligned}
-\left(g_x^{\mathrm{pre}}\right)^\top\delta\theta_0
={}&
\frac{\eta}{|\mathcal S_{\mathrm{tr}}|}
\mathbf{1}[x\in\mathcal S_{\mathrm{tr}}]
\left\|g_x^{\mathrm{pre}}\right\|^2\\
&+
\frac{\eta}{|\mathcal S_{\mathrm{tr}}|}
\sum_{\substack{x'\in\mathcal S_{\mathrm{tr}}\\x'\ne x}}
\left\langle g_x^{\mathrm{pre}},g_{x'}^{\mathrm{pre}}\right\rangle .
\end{aligned}
\end{equation}
The first term is the contribution of $x$'s own training gradient to the
projection; it exists only when $x$ is a member and is always non-negative.
The second term records the cross-prompt effect of the training-member
gradients on the candidate $x$.
For a member candidate, it sums over the other training members; for a
non-member candidate, it sums over all training members because $x$ is absent
from $\mathcal S_{\mathrm{tr}}$.
Each inner product can be positive or negative because the two prompt gradients
may point in the same or opposite directions.
When this cross-prompt term is comparable across the two groups, the additional
non-negative term makes larger member projections more likely without ordering
every member--non-member pair.
The controlled comparison below tests whether this tendency remains visible in
the completed oracle projection after the full OPD run.

For matched member and non-member prompts, we compute the gradient-based teacher-directed audit projection in the LoRA parameter space on shared pre-student contexts. Higher values
indicate stronger alignment with the initial teacher-loss descent direction. As shown in Figure~\ref{fig:oracle_teacher_projection}, the member distributions shift toward higher values across Qwen, InternLM, and Gemma, with $p<0.001$ for every comparison.
This result shows that the completed update is more strongly teacher-directed on member prompts.

\mypara{Forward-Only Estimation}
Computing $q_{\mathrm{proj}}^T$ requires student parameters and gradients,
which the auditor cannot access.
We estimate it from model outputs by averaging the corresponding projection
along the straight path between the two checkpoints.
Let
$\theta_\alpha=\theta_{\mathrm{pre}}+\alpha\Delta\theta$ and define
$q_x^T(\alpha)=
-\nabla_\theta
\mathcal D_x^T(\theta_\alpha;\mu_x^{\mathrm{pre}})^\top\Delta\theta$.
This quantity projects the checkpoint displacement onto the local
teacher-discrepancy descent direction at $\theta_\alpha$, with
$q_x^T(0)=q_{\mathrm{proj}}^T(x)$.
Along this path, we keep $\mu_x^{\mathrm{pre}}$ fixed so that the endpoint
difference compares the two policies at the same inputs.
The fundamental theorem of calculus then gives
\begin{equation}
\label{eq:path_average}
\begin{aligned}
&\mathcal D_x^T(\theta_{\mathrm{pre}};\mu_x^{\mathrm{pre}})
-\mathcal D_x^T(\theta_{\mathrm{post}};\mu_x^{\mathrm{pre}})\\
&\qquad
=-\int_0^1
\nabla_\theta
\mathcal D_x^T(\theta_\alpha;\mu_x^{\mathrm{pre}})^\top
\Delta\theta\,\mathrm d\alpha
=\int_0^1 q_x^T(\alpha)\,\mathrm d\alpha .
\end{aligned}
\end{equation}
The endpoint reduction is therefore the exact path average of the
teacher-directed projection on the fixed pre-student contexts.
It requires only forward evaluations at the two endpoints, which gives
Teacher Alignment Gain (TAG):
\begin{equation}
\label{eq:tag}
\begin{aligned}
s_{\mathrm{TAG}}(x)
&=\mathcal D_x^T(\theta_{\mathrm{pre}};\mu_x^{\mathrm{pre}})
-\mathcal D_x^T(\theta_{\mathrm{post}};\mu_x^{\mathrm{pre}})\\
&=\mathcal A_x^{\mathrm{pre}}\!\left[
\KL\!\left(\pi_{\mathrm{pre}}\Vert\pi_T\right)
-\KL\!\left(\pi_{\mathrm{post}}\Vert\pi_T\right)\right].
\end{aligned}
\end{equation}
A positive TAG means that the post-distillation student is closer to the
teacher on the fixed pre-student contexts.
Candidate and reference prompts are scored under this same convention before
feature standardization.
The evaluation tests whether this path average preserves the prompt-level
ordering and direction of its starting value $q_{\mathrm{proj}}^T$.

\mypara{Entropy Alignment}
\label{sec:taeg}
Distribution matching can also change the concentration of the student's
predictive distribution.
We use reduction in the teacher--student entropy gap as an empirical companion
to TAG.
For any policy $\pi_j$, define
\begin{equation}
\label{eq:model_entropy}
H_j(c_t)=
-\sum_{v\in\mathcal V}\pi_j(v\mid c_t)\log\pi_j(v\mid c_t).
\end{equation}
At the endpoints, $j\in\{\mathrm{pre},\mathrm{post},T\}$.
The aggregate uncertainty gap is
$\mathcal U_x^T(\theta;\mu_x^{\mathrm{pre}})
=\mathcal A_x^{\mathrm{pre}}[|H_\theta-H_T|]$.
Its endpoint reduction is
\begin{equation}
\label{eq:taeg}
s_{\mathrm{TAEG}}(x)
=\mathcal U_x^T(\theta_{\mathrm{pre}};\mu_x^{\mathrm{pre}})
-\mathcal U_x^T(\theta_{\mathrm{post}};\mu_x^{\mathrm{pre}}).
\end{equation}
We call this auxiliary statistic Teacher-Aligned Entropy Gain (TAEG).
A positive TAEG means that the post-distillation student's uncertainty is
closer to the teacher's on the audit contexts.
The audit classifier uses this uncertainty-gap reduction as a separate
coordinate alongside TAG.

\subsection{Access-Aware Signal Computation}
\label{sec:access_estimation}

To apply the teacher-referenced statistics under different access budgets, we
adapt only the representation of the teacher distribution.
Full-distribution access uses $\pi_T(\cdot\mid c_t)$ directly, while SD remains
unchanged in every setting because both student checkpoints are available.

Under Top-$K$ access, grouping the hidden probability mass gives all three
models a common support.
Let $\Omega_{K,t}(x)\subseteq\mathcal V$ be the $K$ tokens with the highest
teacher probabilities at context $c_t$.
For $j\in\{\mathrm{pre},\mathrm{post},T\}$, we omit the common condition
$c_t$ and define
\begin{equation}
\label{eq:grouped_distribution}
\pi_{j,\mathrm{grp}}^K(u)=
\begin{cases}
\pi_j(u), & u\in\Omega_{K,t}(x),\\
1-\sum_{v\in\Omega_{K,t}(x)}\pi_j(v), & u=\bot.
\end{cases}
\end{equation}
TAG computes KL divergence on
$\Omega_{K,t}(x)\cup\{\bot\}$, preserving the teacher's visible probabilities
and total residual mass.
For the entropy feature, define
$H_{j,\mathrm{grp}}^K=-\sum_u\pi_{j,\mathrm{grp}}^K(u)
\log\pi_{j,\mathrm{grp}}^K(u)$ and use the same endpoint gap reduction as
Eq.~\ref{eq:taeg}.
This top-$K$ entropy feature measures concentration across the visible teacher
tokens and the residual bucket.
We denote it by $s_{\mathrm{TAEG}}^K$ and use it as the top-$K$ counterpart of
full-distribution TAEG.
To quantify how much student probability remains individually resolved, we
also compute
$\rho_K(x)=\mathcal A_x^{\mathrm{pre}}
[\sum_{v\in\Omega_{K,t}(x)}\pi_{\mathrm{post}}(v)]$.

When the true teacher is unavailable, a surrogate $\widetilde T$ supplies the
reference distribution in TAG and TAEG.
The resulting signals measure checkpoint progress relative to this accessible
teacher reference.

\subsection{Evidence Aggregation}
\label{sec:membership_prediction}

The augmented predictor adds checkpoint-calibrated Zlib~\cite{carlini2021extracting} and Min-K~\cite{shi2023detecting} as generic confidence controls.
Both compare the post- and pre-distillation checkpoints on the sampled responses.
Calibrated Zlib divides the generated-token log-likelihood-ratio sum by the compressed response length, while calibrated Min-K takes the difference between the two checkpoints' mean log probabilities over their respective lowest-$5\%$ token positions.
We organize the audit statistics into nested feature vectors:
\begin{equation}
\label{eq:pama_signals}
\boldsymbol{\phi}(x) =[s_{\mathrm{Zlib}}^{\mathrm{cal}},s_{\mathrm{MinK}}^{\mathrm{cal}},
s_{\mathrm{SD}},s_{\mathrm{TAG}},s_{\mathrm{TAEG}}]^\top.
\end{equation}
Under Top-$K$ access, $s_{\mathrm{TAEG}}$ in these vectors denotes the grouped counterpart $s_{\mathrm{TAEG}}^K$.
All components are evaluated at $x$, with the argument omitted for readability.

We standardize the selected features and fit a logistic regressor using only each audit-training fold.
A larger output indicates stronger evidence that $x$ is an OPD member.
We report repeated stratified five-fold out-of-fold scores.

\section{Experiments}

\subsection{Experiment Setup}

\mypara{Datasets}
We evaluate \sysname on six datasets covering diverse tasks:
MATH~\cite{hendrycks2021math} focuses on mathematical problem solving, while MBPP~\cite{austin2021program} targets code generation from natural-language specifications.
Toy-add is a synthetic boundary dataset consisting of simple arithmetic prompts.
MulDimIF~\cite{ye2025muldimif} for multi-dimensional instruction following, the STEM subset of MMLU-Pro~\cite{wang2024mmlupro} for cross-domain STEM reasoning, and BigCodeBench Instruct~\cite{zhuo2025bigcodebench} for complex code generation are also included.

\mypara{Models}
We consider three instruction-tuned teacher-student model families:
Qwen2.5-3B-Instruct and Qwen2.5-14B-Instruct (Qwen), InternLM2.5-1.8B-Chat and InternLM2.5-7B-Chat (InternLM), and Gemma-3-4B-IT and Gemma-3-12B-IT (Gemma).
For each family, the smaller model serves as the student and the larger model serves as the teacher.
The teacher model is fixed throughout the distillation process.

\mypara{Baselines}
We compare \sysname with representative membership auditing methods, including Likelihood (Like)~\cite{duan2024membership}, Zlib~\cite{carlini2021extracting}, Min-K~\cite{shi2023detecting}, and DIBA~\cite{liu2026diba}.
For Min-K, we follow the original implementation and use the lowest-$5\%$ token log-probabilities.
For a fair comparison under OPD, we adopt checkpoint-calibrated variants of Zlib and Min-K, where their scores are calibrated using the pre-distillation student model $S_{\mathrm{pre}}$.
For DIBA, we only retain its policy-shift component, since other signals require RL-specific training information that is unavailable in OPD.

For \sysname, we randomly sample 1000 member and 1000 non-member prompts from each dataset to train the logistic regression classifier.
The classifier is evaluated with five independent runs using stratified five-fold cross-validation, and the averaged score is reported as the final auditing score.

During auditing, 1000 member and 1000 non-member prompts are sampled with matched distributions.
We set the sampling temperature to 0.8.
For each prompt, we generate $R=10$ responses with a maximum length of 128 tokens.
We report the maximum TPR achieved at an empirical FPR no larger than $0.1\%$.
We set $K=10$ under Top-$K$ access scenario.

\mypara{Metrics}
Following prior membership auditing studies~\cite{duan2024membership, liu2026diba}, we use ROC-AUC and True Positive Rate (TPR) under low false positive rates.
Specifically, we report TPR@0.1\%FPR, TPR@0.5\%FPR, and TPR@1\%FPR.
The strictest setting, TPR@0.1\%FPR, is emphasized because practical auditing scenarios usually require a highly constrained false-positive budget.

\mypara{Other Settings}
All experiments are conducted on NVIDIA A800 80GB GPUs.
For OPD training, we use member prompts with LoRA fine-tuning, where the LoRA rank and alpha are set to 32 and 64, respectively.
The learning rate is $1\times10^{-4}$, and all prompts are formatted using the corresponding chat templates.

\begin{figure}[t]
    \centering
    \includegraphics[width=\columnwidth]{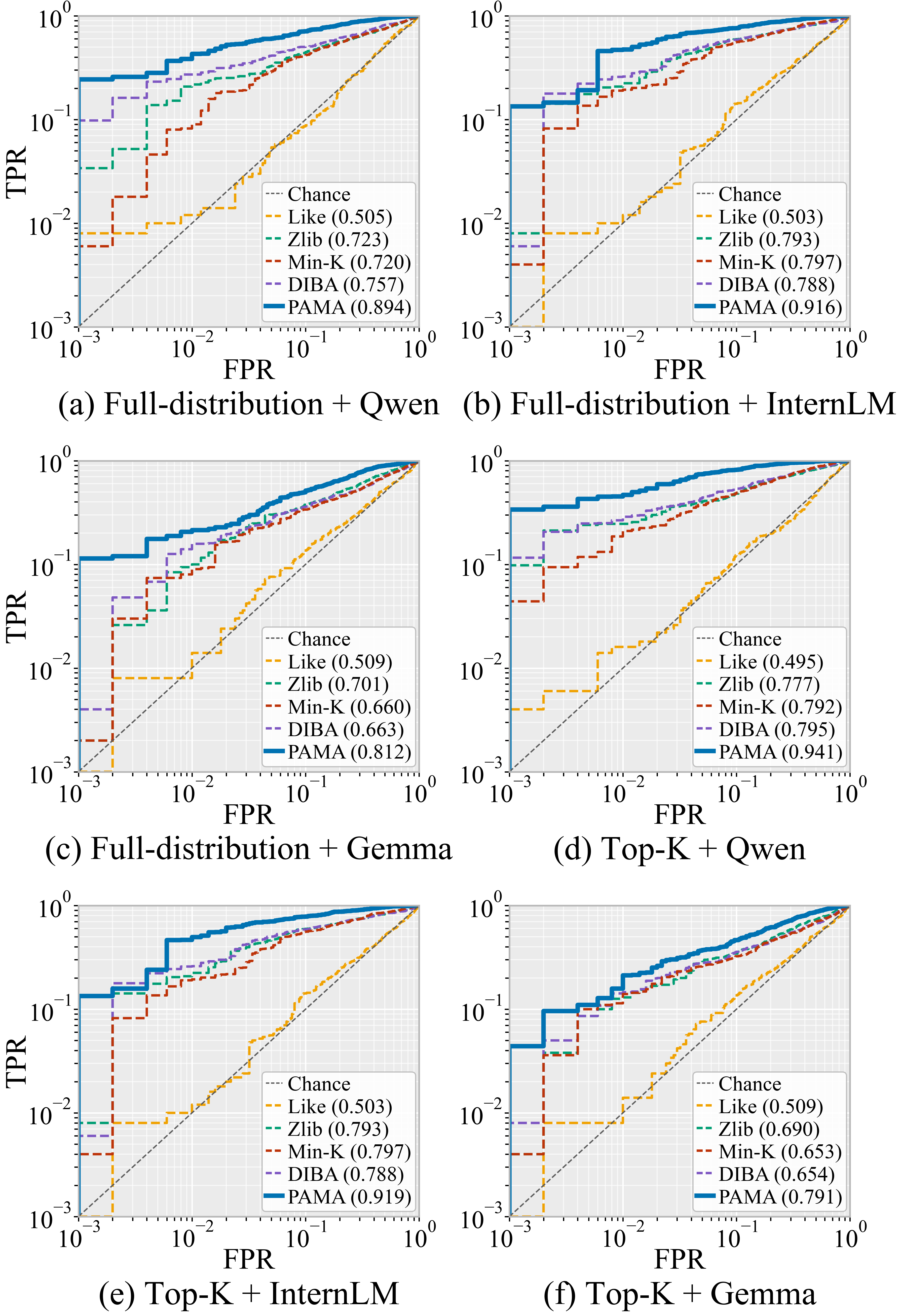}
    \caption{ROC curves on the MATH dataset under full-distribution and Top-$K$ access, with ROC-AUC scores reported in parentheses in the legend.}
    \label{fig:roc-math-all}
\end{figure}

\subsection{Main Auditing Results}

\mypara{Prerequisite: Effective OPD}
Subsequent experiments rely on the premise that OPD effectively transfers knowledge from the teacher model to the student model.
We therefore first compare the student model’s task accuracy before and after OPD.
On the MATH dataset, OPD improves the accuracy from 65.3\% to 84.3\% for Qwen, from 43.2\% to 67.7\% for the InternLM family, and from 52.1\% to 78.9\% for the Gemma family.
These results confirm the effectiveness of OPD and establish the premise for the following auditing experiments.

\mypara{Full-distribution and Top-$K$ Access}
Figure~\ref{fig:roc-math-all} presents the ROC curves of \sysname and the evaluated baselines on MATH across three model families, under both full-distribution and Top-K teacher access. The AUC values reported in parentheses show that \sysname consistently achieves the strongest overall performance in all six settings, while maintaining a clear advantage in the low-FPR region. Interestingly, Top-K access matches or outperforms full-distribution access for Qwen and InternLM, although a small decrease is observed for Gemma. This result suggests that the membership-relevant alignment signal is mainly concentrated on the teacher’s high-probability tokens, whereas the long tail of low-probability entries may introduce noise into distributional distance estimation. Therefore, restricted teacher access is not necessarily a weaker approximation: retaining only the most informative probability mass can also serve as an effective denoising mechanism.

\subsection{Further Understanding}

\begin{figure}[t]
    \centering
    \includegraphics[width=\columnwidth]{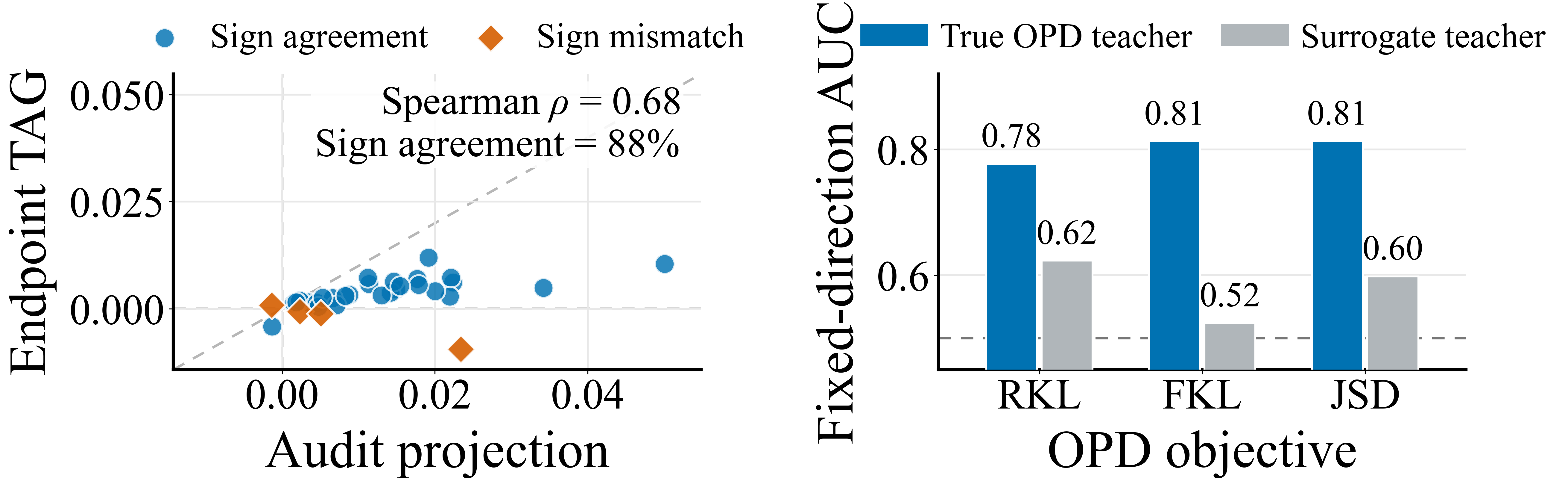}
    \caption{Validation of TAG. Left: agreement with the gradient-based audit projection. Right: fixed-direction AUC across OPD objectives and teacher references.}
    \label{fig:teacher_direction_validation}
\end{figure}

\mypara{Validation of TAG Signal}
Figure~\ref{fig:teacher_direction_validation} compares forward-only TAG with the gradient-based audit projection and examines the effect of the teacher reference. TAG agrees with the projection direction on 88\% of prompts, showing that endpoint evaluations preserve
the teacher-directed update signal without gradient access.
Across different OPD objectives, using the true OPD teacher consistently achieves better auditing performance than using a surrogate teacher. This result confirms that TAG is most reliable when its reference matches the teacher used for distillation.

\begin{table}[t]
\centering
\small
\setlength{\tabcolsep}{3.2pt}
\begin{tabular}{lcccc}
\toprule
Datasets & AUC & TPR@0.1\% & TPR@0.5\% & TPR@1\% \\

\midrule
\multicolumn{5}{c}{\textbf{Full-distribution Access}}\\
\midrule

MMLU-Pro  & 0.810 & 0.066 & 0.110 & 0.132 \\
BigCodeBench & 0.821 & 0.076 & 0.133 & 0.173 \\
MulDimIF & 0.813 & 0.058 & 0.122 & 0.206 \\

\midrule
\multicolumn{5}{c}{\textbf{Top-$K$ Access}}\\
\midrule

MMLU-Pro  & 0.872 &0.064 & 0.134 & 0.380 \\
BigCodeBench &0.769  & 0.066 & 0.101 & 0.262 \\
MulDimIF &0.773  &0.050  &0.106  &0.302  \\



\bottomrule
\end{tabular}
\caption{Audit performance of \sysname across datasets using the Qwen family.}
\label{tab:different-datasets}
\end{table}

\mypara{Different Datasets on Qwen}
Table ~\ref{tab:different-datasets} reports the auditing performance of \sysname on MMLU-Pro, BigCodeBench, and MulDimIF using the Qwen teacher–student family.
\sysname maintains clearly above-chance AUC across reasoning, code-generation, and instruction-following tasks, demonstrating that its effectiveness is not limited to the MATH benchmark. The relative behavior of the two access settings is nevertheless task dependent: Top-K access provides a clear benefit on MMLU-Pro, while full-distribution access produces stronger overall ranking on BigCodeBench and MulDimIF. Moreover, changes in AUC do not always coincide with changes in low-FPR TPR, indicating that global ranking quality and reliable detection under a strict false-positive budget capture different aspects of auditability. These results motivate evaluating membership auditing with both AUC and low-FPR metrics rather than relying on either measure alone.

\begin{table}[t]
\centering
\small
\setlength{\tabcolsep}{2pt}
\begin{tabular}{llcccc}
\toprule
Model & Signals
& AUC & TPR@0.1\% & TPR@0.5\% & TPR@1\% \\
\midrule

\multirow{4}{*}{Qwen}
& Trad.
& 0.727 & 0.016 & 0.104 & 0.168 \\
& +SD
& 0.772 & 0.044 & 0.196 & 0.240 \\
& +TAG
& 0.879 & 0.186 & 0.238 & 0.330 \\
& +TAEG
& \textbf{0.894} & \textbf{0.244} & \textbf{0.418} & \textbf{0.438} \\
\midrule

\multirow{4}{*}{InternLM}
& Trad.
& 0.801 & 0.002 & 0.168 & 0.204 \\
& +SD
& 0.803 & 0.002 & 0.176 & 0.248 \\
& +TAG
& 0.905 & 0.102 & 0.190 & 0.394 \\
& +TAEG
& \textbf{0.916} & \textbf{0.134} & \textbf{0.192} & \textbf{0.474} \\
\midrule

\multirow{4}{*}{Gemma}
& Trad.
& 0.729 & 0.000 & 0.024 & 0.076 \\
& +SD
& 0.742 & 0.004 & 0.034 & 0.098 \\
& +TAG
& 0.803 & 0.072 & 0.164 & 0.184 \\
& +TAEG
& \textbf{0.812} & \textbf{0.114} & \textbf{0.176} & \textbf{0.214} \\
\bottomrule
\end{tabular}%
\caption{Cumulative signal ablation on MATH under full-distribution access.}
\label{tab:different_signal}
\end{table}

\mypara{Ablation Study of Different Signals}
Table~\ref{tab:different_signal} presents a cumulative ablation of the signals used by \sysname, starting from the calibrated confidence features and sequentially introducing SD, TAG, and TAEG. Adding these signals progressively improves the final auditing performance across all three model families, with the complete combination achieving the strongest results. SD provides a useful but generally modest improvement, showing that the magnitude of student policy change alone contains limited membership information. In contrast, introducing TAG produces the largest performance gain, particularly under strict false-positive constraints, confirming that the teacher-directed component of the update is the primary signal missing from conventional policy-drift auditing. TAEG further improves the final result by capturing complementary changes in predictive uncertainty.The ablation therefore supports the central design of \sysname: student drift describes how much the policy changes, whereas TAG and TAEG reveal whether that change follows the teacher.

\begin{table}[t]
\centering
\small
\setlength{\tabcolsep}{5pt}
\begin{tabular}{@{}lccc@{}}
\toprule
Setting & AUC & TPR@0.1\% & TPR@1\% \\
\midrule
FKL / Full       & 0.812 & 0.086  & 0.202 \\
FKL / Top-$K$    & 0.865 & 0.138 & 0.302 \\
JSD / Full       & 0.780 & 0.091  & 0.258 \\
JSD / Top-$K$    & 0.824 & 0.152 & 0.234 \\
\bottomrule
\end{tabular}
\caption{Effect of OPD objectives and teacher access on Qwen-MATH auditing.}
\label{tab:opd-algorithm}
\end{table}

\mypara{Impact of Different OPD Algorithms}
Table~\ref{tab:opd-algorithm} evaluates \sysname on Qwen-MATH under two representative OPD objectives: forward KL divergence (FKL), which uses an asymmetric KL objective, and Jensen--Shannon divergence (JSD), which symmetrically compares the student and teacher distributions.
Their formulations and implementation details are provided in the Appendix. 
\sysname remains effective across both objectives and access settings, showing that its alignment signals are not tied to a specific OPD algorithm. Top-$K$ access achieves higher AUC for both objectives, indicating that the teacher's high-probability outputs retain most membership-relevant alignment information. Performance at low-FPR operating points varies across objectives, suggesting that the OPD loss changes how teacher-guided updates are distributed across prompts without eliminating the resulting membership trace.

\begin{figure}[t]
    \centering
    \includegraphics[width=\columnwidth]{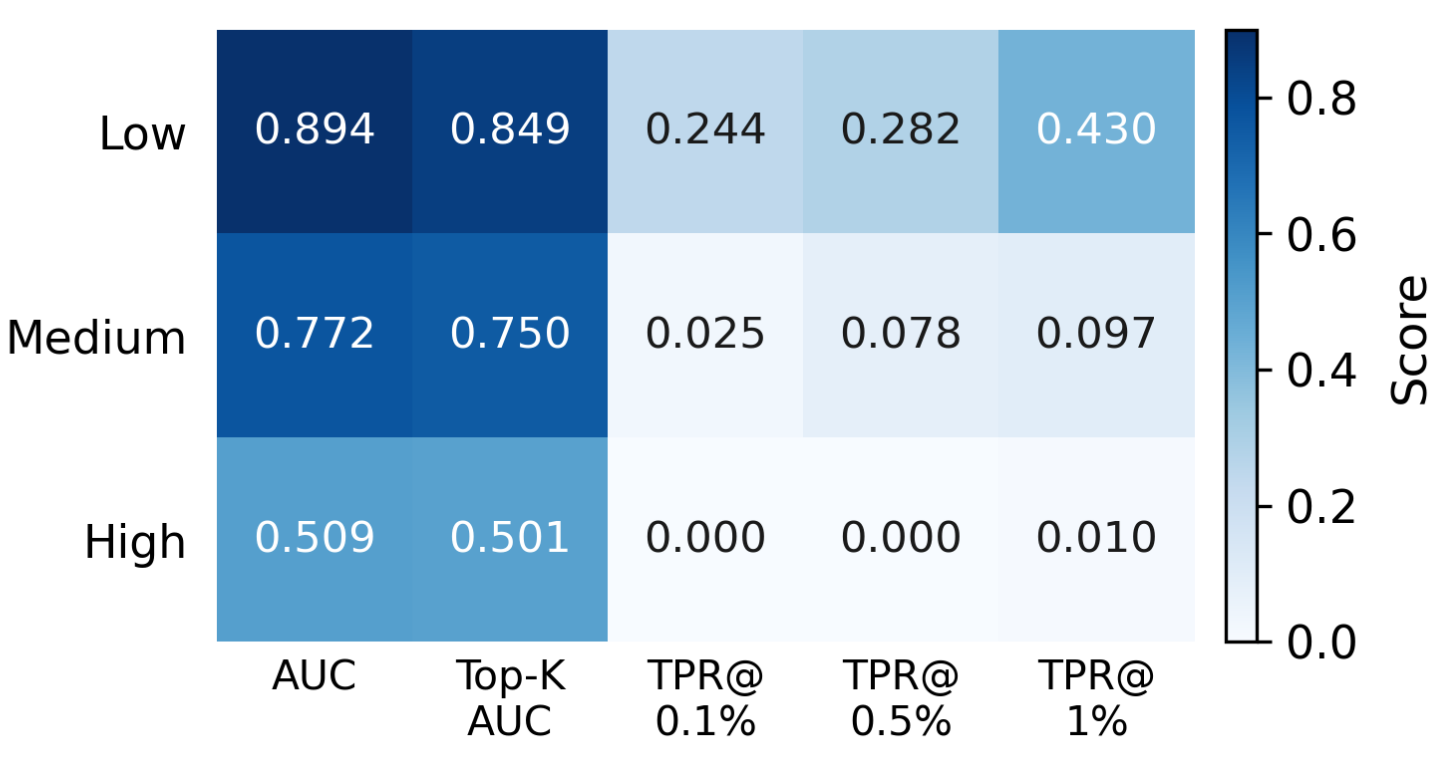}
    \caption{Effect of cross-prompt generalization on auditability.}
    \label{fig:auditability-boundaries}
\end{figure}

\mypara{Impact of Prompt Generalization Level}
Figure~\ref{fig:auditability-boundaries} compares three task constructions with different levels of cross-prompt generalization, where MATH, MBPP, and Toy represent high, medium, and low levels, respectively.
Auditability varies substantially across these settings: limited cross-prompt transfer yields strong AUC and reliable low-FPR detection, moderate transfer weakens strict-FPR detection, and high generalization approaches random guessing.
When prompts share similar solution patterns, training on one prompt can also improve teacher alignment on related non-members, obscuring the distinction between direct participation and indirect generalization.
This result identifies cross-prompt generalization as a fundamental boundary of membership auditing: \sysname is effective when prompt-specific training traces are not overwhelmed by transferable updates.

\begin{figure}[t]
    \centering
    \begin{minipage}[t]{0.46\columnwidth}
        \centering
        \includegraphics[width=\linewidth]{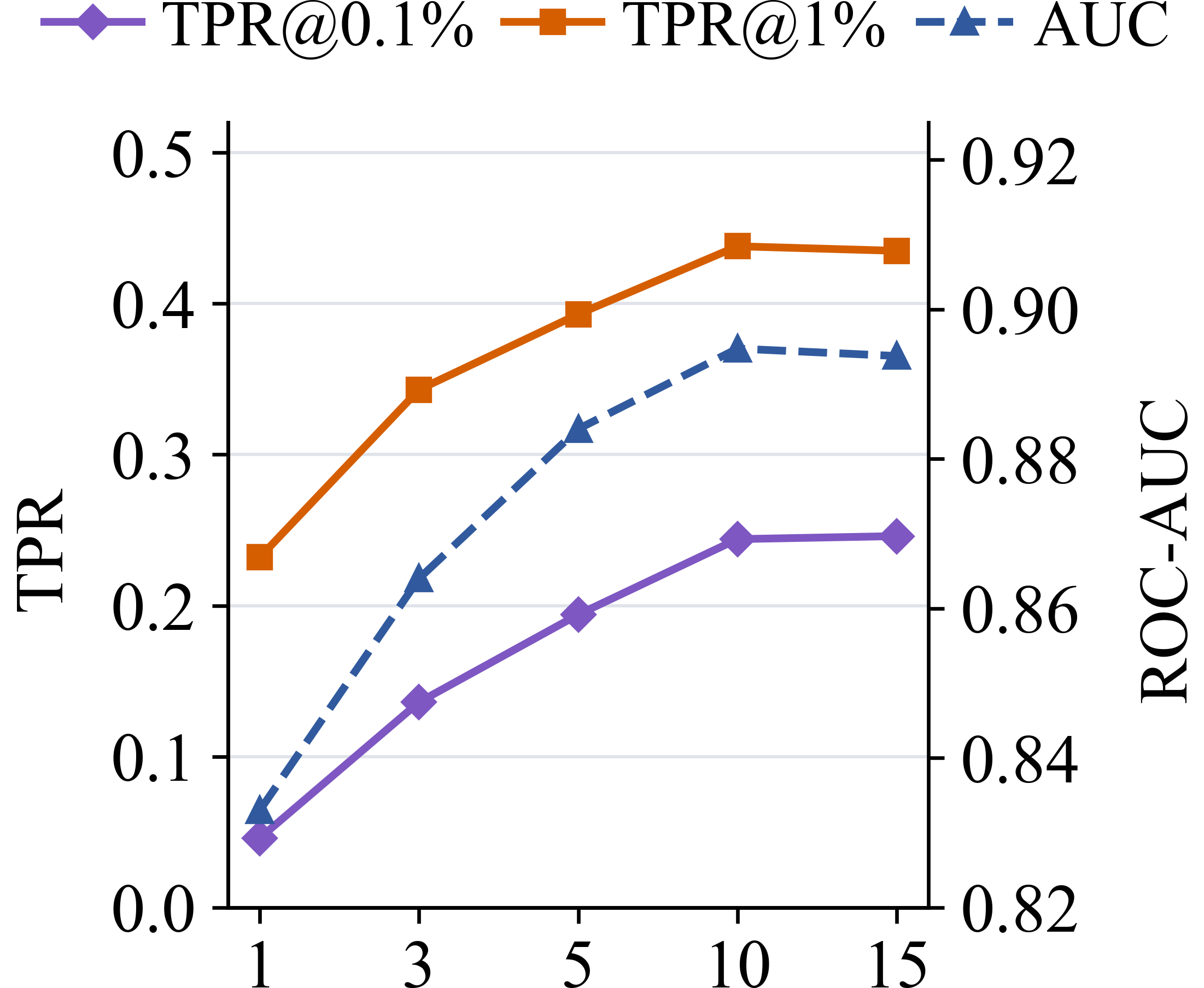}
        \caption{Sensitivity to $K$ on Qwen-MATH under Top-$K$ access.}
        \label{fig:sensitivity_topk}
    \end{minipage}
    \hfill
    \begin{minipage}[t]{0.46\columnwidth}
        \centering
        \includegraphics[width=\linewidth]{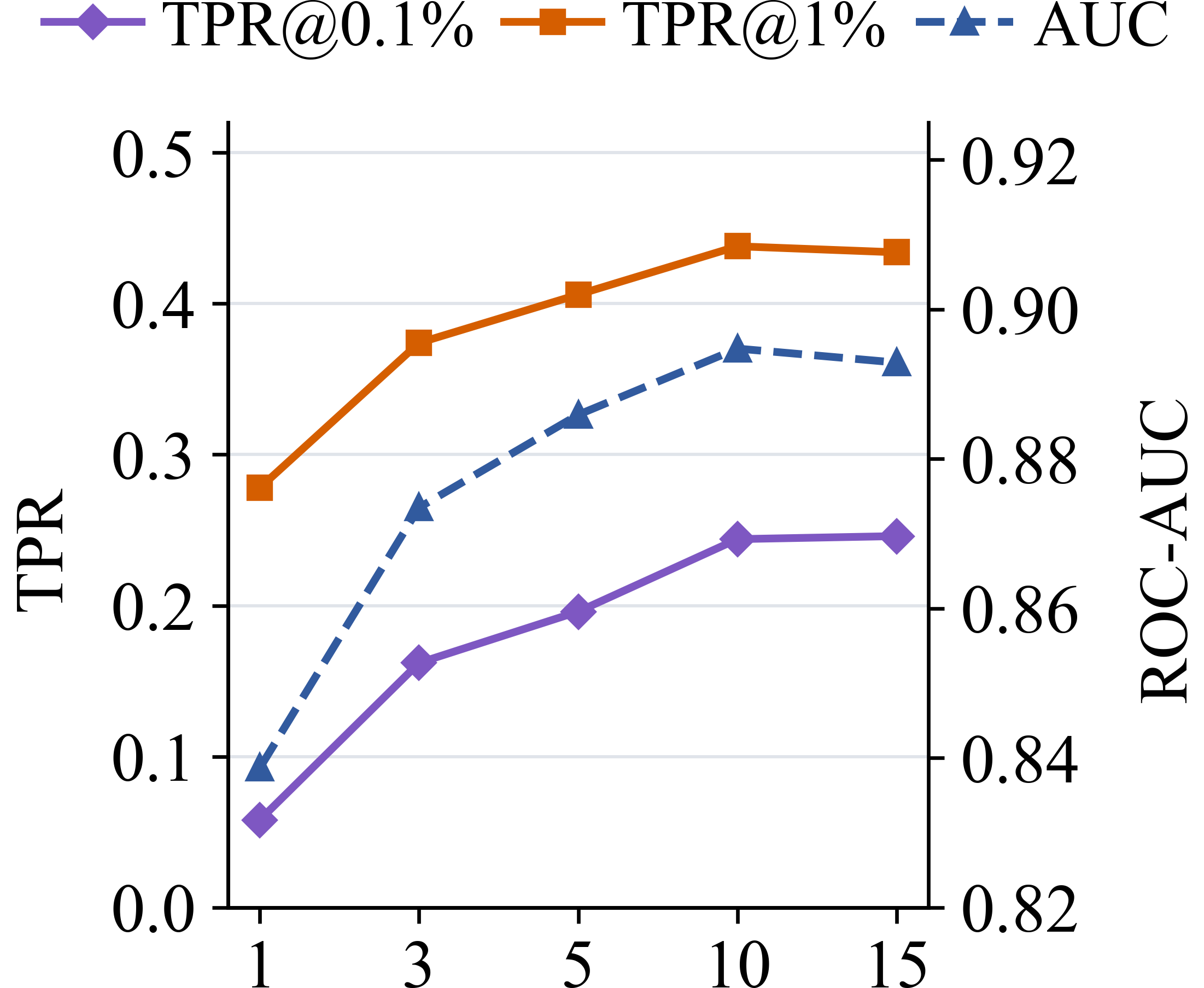}
        \caption{Sensitivity to $R$ on Qwen-MATH under Full-distribution access.}
        \label{fig:sensitivity_response_count}
    \end{minipage}
\end{figure}

\mypara{Impact of $K$}
Figure~\ref{fig:sensitivity_topk} studies the effect of the teacher-access budget K under Top-K access. Increasing K initially improves both AUC and low-FPR detection, with most of the gain obtained by K=10. Performance then largely saturates, and exposing additional teacher probabilities at K=15 provides little further benefit. This trend indicates that a small set of high-probability teacher tokens is sufficient to capture most of the teacher-aligned update signal, while probabilities deeper in the distribution tail contribute limited additional information. We therefore use K=10 as the default setting, balancing auditing performance against the amount of teacher information required.

\mypara{Impact of Response Count}
The response count $R$ denotes the number of independently sampled responses per candidate prompt.
We average the auditing signals over these trajectories before membership prediction.
Figure~\ref{fig:sensitivity_response_count} shows that increasing $R$ improves performance when few responses are available, but the gains diminish as $R$ approaches 10.
This is because additional responses provide more representative audit contexts and reduce randomness from on-policy generation.
Beyond $R=10$, performance changes little while query and generation costs continue to increase. We therefore use $R=10$ in the main experiments.



\begin{figure}[t]
\centering
\includegraphics[width=\columnwidth]{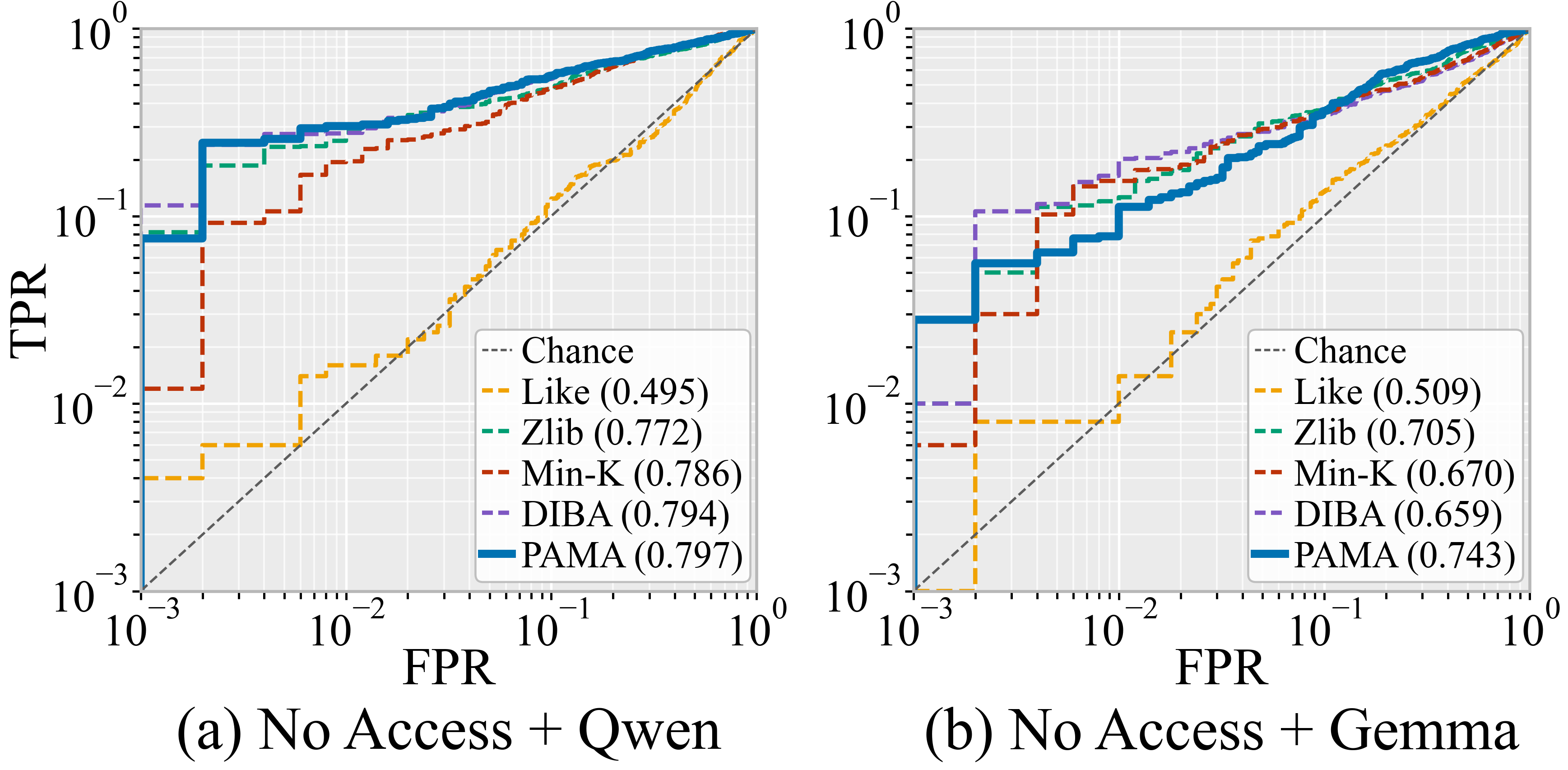}     
\caption{ROC curves under surrogate-teacher access without true OPD teacher model.}
\label{fig:roc-black-all}
\end{figure}

\mypara{No True-Teacher Access}
Figure~\ref{fig:roc-black-all} evaluates auditing without access to the original OPD teacher. We use a compatible general-purpose surrogate from the same model family and generation as the student, with a matching tokenizer, vocabulary, and prompt format but different parameters from the original teacher.
Therefore, \sysname remains comparable performances on Gemma and Qwen, showing that useful teacher-alignment evidence can still be recovered.
Its advantage over policy-drift auditing is nevertheless reduced because the surrogate induces a different output distribution and thus provides only an approximate alignment direction.
Overall, a compatible surrogate serves as a practical fallback, although auditing quality depends on how closely it matches the unavailable teacher.

\section{Conclusion}
In this work, we study prompt-level membership auditing in OPD and identify teacher-directed policy alignment as a new membership signal. Based on the observation that member prompts directly contribute to teacher-guided updates, we propose \sysname, which combines Student Drift, Teacher Alignment Gain, Teacher-Aligned Entropy Gain, and calibrated confidence signals. \sysname requires only forward evaluations of the pre- and post-distillation students and therefore avoids access to training gradients or trajectories. Experiments across multiple datasets, model families, OPD objectives, and teacher-access settings demonstrate that \sysname consistently improves auditing performance over existing likelihood- and drift-based methods. Further analyses show that high-probability teacher outputs contain most of the useful alignment information, while strong cross-prompt generalization can weaken membership distinguishability. These results establish teacher-referenced policy alignment as an effective signal for auditing data usage in OPD.

\bibliography{aaai2027}

\clearpage
\maketitlesupplementary

\appendix

\section{Implementation Details of OPD Variant Algorithms}
\label{app:opd-variants}

We evaluate three OPD objectives that differ only in the
distribution-alignment loss used on student-generated trajectories.
For each member prompt $x$, the current student samples a response
$y \sim \pi_{\theta}(\cdot \mid x)$. At each token position $t$, both
the student and teacher distributions are evaluated under the same
context $c_t=(x,y_{<t})$. The token-level reverse-KL objective used in
the default OPD setting is
\begin{equation}
\ell_{\mathrm{RKL}}(c_t)
=
D_{\mathrm{KL}}
\left(
\pi_{\theta}(\cdot \mid c_t)
\;\middle\|\;
\pi_T(\cdot \mid c_t)
\right).
\end{equation}
The forward-KL variant reverses the two distributions:
\begin{equation}
\ell_{\mathrm{FKL}}(c_t)
=
D_{\mathrm{KL}}
\left(
\pi_T(\cdot \mid c_t)
\;\middle\|\;
\pi_{\theta}(\cdot \mid c_t)
\right).
\end{equation}
Because the expectation is taken under the teacher distribution, FKL
places greater emphasis on tokens assigned high probability by the
teacher and penalizes the student for failing to cover these regions.

For Jensen--Shannon divergence, we first define
\begin{equation}
m_t
=
\frac{1}{2}
\left(
\pi_{\theta}(\cdot \mid c_t)
+
\pi_T(\cdot \mid c_t)
\right),
\end{equation}
and compute
\begin{equation}
\begin{aligned}
\ell_{\mathrm{JSD}}(c_t)
=&
\frac{1}{2}
D_{\mathrm{KL}}
\left(
\pi_{\theta}(\cdot \mid c_t)
\;\middle\|\;
m_t
\right)\\
&\qquad
+
\frac{1}{2}
D_{\mathrm{KL}}
\left(
\pi_T(\cdot \mid c_t)
\;\middle\|\;
m_t
\right).
\end{aligned}
\end{equation}
Unlike the asymmetric KL objectives, JSD symmetrically constrains the
discrepancy between the student and teacher distributions.

For each variant, the final OPD loss is obtained by averaging the
corresponding token-level loss over all positions of the sampled
response:
\begin{equation}
\mathcal{L}_{\mathrm{OPD}}
=
\mathbb{E}_{x,\,y\sim\pi_{\theta}(\cdot\mid x)}
\left[
\frac{1}{L(y)}
\sum_{t=1}^{L(y)}
\ell(c_t)
\right],
\end{equation}
where $\ell$ is instantiated as $\ell_{\mathrm{RKL}}$,
$\ell_{\mathrm{FKL}}$, or $\ell_{\mathrm{JSD}}$.

\paragraph{Training Configuration.}
All variants use the same 1,000 member prompts and are trained for
10 epochs. We update the student using LoRA with rank $r=32$ and
scaling parameter $\alpha=64$. The learning rate is set to
$1\times10^{-4}$. During each OPD step, the current student generates
a response with temperature $0.8$, top-$p$ sampling with
$p=0.95$, and at most 128 new tokens. The teacher distribution is
then evaluated along this student-generated trajectory to compute the
distillation loss. RKL, FKL, and JSD differ only in the alignment loss;
the training prompts, teacher--student models, LoRA configuration,
number of epochs, and generation parameters are otherwise identical.

\section{Implementation Details of Logistic Regressor}
\label{app:logistic-regressor}

For each candidate prompt \(x\), we sample \(R=10\) responses from
\(S_{\mathrm{pre}}\) for Zlib, Min-K, TAG, and TAEG, and independently
sample \(R=10\) responses from \(S_{\mathrm{post}}\) for SD. Within each
signal, the pre-distillation student, post-distillation student, and
teacher are evaluated on the same realized response contexts. The five
audit signals are aggregated into the feature vector
\begin{equation}
\begin{aligned}
\boldsymbol{\phi}(x)
={}&
\bigl[
s_{\mathrm{Zlib}}^{\mathrm{cal}}(x),
s_{\mathrm{MinK}}^{\mathrm{cal}}(x),
s_{\mathrm{SD}}(x), \\
&\qquad
s_{\mathrm{TAG}}(x),
s_{\mathrm{TAEG}}(x)
\bigr]^{\top}
\in \mathbb{R}^{5}.
\end{aligned}
\end{equation}

The labeled reference set contains 1,000 member prompts and
1,000 matched non-member prompts, producing a feature matrix
$\mathbf{X}\in\mathbb{R}^{2000\times 5}$ and binary labels
$m_i\in\{0,1\}$.

\paragraph{Training Procedure.}
We use repeated stratified five-fold cross-validation to train the
logistic regressor. In each split, four folds containing 800 members
and 800 non-members are used for training, while the remaining fold
containing 200 members and 200 non-members is held out for
prediction. Feature standardization is performed independently
within each split. Specifically, for each feature coordinate $j$, we
compute the mean $\mu_j$ and standard deviation $\sigma_j$ using
only the 1,600 training samples:
\begin{equation}
\widetilde{\phi}_j(x)
=
\frac{\phi_j(x)-\mu_j}{\sigma_j}.
\end{equation}
The same training-fold statistics are used to standardize the
held-out samples, preventing information leakage from the held-out
fold.

Given a standardized feature vector
$\widetilde{\boldsymbol{\phi}}(x)$, the logistic regressor computes
\begin{equation}
z(x)
=
\mathbf{w}^{\top}
\widetilde{\boldsymbol{\phi}}(x)+b,
\end{equation}
where $\mathbf{w}\in\mathbb{R}^{5}$ contains the weights of the five
audit signals and $b$ is the intercept. The corresponding membership
probability is
\begin{equation}
p(x)
=
\sigma(z(x))
=
\frac{1}{1+\exp(-z(x))}.
\end{equation}
The parameters are optimized using binary cross-entropy with
$\ell_2$ regularization:

\begin{equation}
\begin{aligned}
\mathcal{L}_{\mathrm{LR}}
=&
-\sum_{i\in\mathcal{I}_{\mathrm{train}}}
\Big[
m_i\log p(x_i) \\
&\qquad
+(1-m_i)\log\bigl(1-p(x_i)\bigr)
\Big]
+\lambda\lVert\mathbf{w}\rVert_2^2 .
\end{aligned}
\label{eq:lr-loss}
\end{equation}

We use an $\ell_2$-regularized logistic regressor with inverse
regularization strength $C=1$, the L-BFGS optimizer, an intercept
term, balanced class weights, and at most 5,000 optimization
iterations. Since each training split contains equal numbers of
members and non-members, balanced class weighting assigns identical
weights to the two classes.

\paragraph{Prediction Procedure.}
For each split, the regressor trained on the 1,600 training samples
predicts membership probabilities for the 400 held-out samples.
Repeating this procedure over all five folds produces one
out-of-fold probability for every prompt, where each probability is
generated by a model that was not trained on that prompt.

We repeat the complete stratified five-fold procedure five times
using random seeds
$\{42,1042,2042,3042,4042\}$. Each prompt therefore obtains five repeated out-of-fold probability estimates. Its final auditing score is
computed as
\begin{equation}
\overline{p}(x_i)
=
\frac{1}{5}
\sum_{q=1}^{5}
p_{\mathrm{OOF}}^{(q)}(x_i).
\end{equation}
ROC-AUC and TPR at the specified FPR operating points are computed
from these averaged scores. No fixed classification threshold is
used when computing these metrics.

\end{document}